\documentclass[letterpaper]{article} 
\usepackage{aaai24}  
\usepackage{times}  
\usepackage{helvet}  
\usepackage{courier}  
\usepackage[hyphens]{url}  
\usepackage{graphicx} 
\usepackage{natbib}  
\usepackage{caption} 
\usepackage{algorithm}
\usepackage{algorithmic}
\usepackage{times}
\usepackage{epsfig}
\usepackage{graphicx}

\usepackage{amssymb}
\usepackage{caption}
\usepackage{bbding}
\usepackage{multirow}
\usepackage{subcaption}
\usepackage{tabularx}
\usepackage{booktabs}
\usepackage[autostyle=true]{csquotes}
\usepackage[dvipsnames]{xcolor}
\usepackage{amsmath}
\usepackage{mathtools}
\usepackage[dvipsnames]{xcolor}
\usepackage{xspace}
\usepackage[dvipsnames]{xcolor}

\newcommand{\yq}[1]{\textcolor{black}{#1}}
\newcommand{\my}[1]{\textcolor{black}{#1}}
\newcommand{\CUT}[1]{}
\newcommand{\modelname}{approach\xspace}
\usepackage{mathtools}

\newcommand{\bx}{\mathbf{x}}

\newcommand{\bz}{\mathbf{z}}

\newcommand{\bI}{\mathbf{I}}
\newcommand{\bepsilon}{{\boldsymbol{\epsilon}}}
\newcommand{\defeq}{\coloneqq}

\newcommand{\encoder}{\mathcal{E}}
\newcommand{\decoder}{\mathcal{D}}
\newcommand{\latent}{\bz}

\usepackage{newfloat}
\usepackage{listings}
\DeclareCaptionStyle{ruled}{labelfont=normalfont,labelsep=colon,strut=off} 
\floatstyle{ruled}
\newfloat{listing}{tb}{lst}{}
\floatname{listing}{Listing}
\title{Follow Your Pose: \\
Pose-Guided Text-to-Video Generation using Pose-Free Videos}
\author {
    Yue Ma$^{1}$\thanks{Equal contributions.}\thanks{Work done during an internship at Tencent AI Lab.} \quad 
    Yingqing He$^{2}$\footnotemark[1]  \quad 
    Xiaodong Cun$^3$ \quad
    Xintao Wang$^3$ \quad 
    Siran Chen$^4$ \quad  \\
    Xiu Li$^{1}$\thanks{Corresponding authors.} \quad 
    Qifeng Chen$^{2}$\footnotemark[3]
}
\affiliations{
    \textsuperscript{\rm 1} Tsinghua Shenzhen International Graduate School, Tsinghua University, Shenzhen, China \\
    \textsuperscript{\rm 2}  The Hong Kong University of Science and Technology, Hong Kong \\
    \textsuperscript{\rm 3}  Tencent AI Lab, Shenzhen, China \\
    \textsuperscript{\rm 4} Shenzhen Institute of Advanced Technology, Chinese Academy of Science, Shenzhen, China \\
    \{y-ma21, lixiu\}@mails.tsinghua.edu.cn, yhebm@connect.ust.hk, \{vinthony,xintao.alpha\}@gmail.com
    Chensiran17@mails.ucas.ac.cn, cqf@ust.hk\\ 
}

\usepackage{bibentry}

\begin{document}

\maketitle

\begin{abstract}
Generating text-editable and pose-controllable character videos have an imperious demand in creating various digital human. 
Nevertheless, this task has been 
restricted by the absence of a comprehensive dataset featuring paired video-pose captions and the generative prior models for videos.
In this work, we design a novel two-stage training scheme that can utilize easily obtained datasets~(i.e., image pose pair and pose-free video) and the pre-trained text-to-image~(T2I) model to obtain the pose-controllable character videos. 
Specifically, in the first stage, only the pose-image pairs are used only for controllable text-to-image generation. We learn a zero-initialized convolutional encoder to encode the pose information. In the second stage, we finetune the motion of the above network via a pose-free video dataset by adding the learnable temporal self-attention and reformed cross-frame self-attention blocks.
Powered by our new designs, our method successfully generates continuously pose-controllable character videos while keeping the concept generation and composition ability from the pre-trained T2I model. The code and models are available on \url{https://follow-your-pose.github.io/}.
\end{abstract}

\begin{figure}
    \centering
    \includegraphics[width=\linewidth]{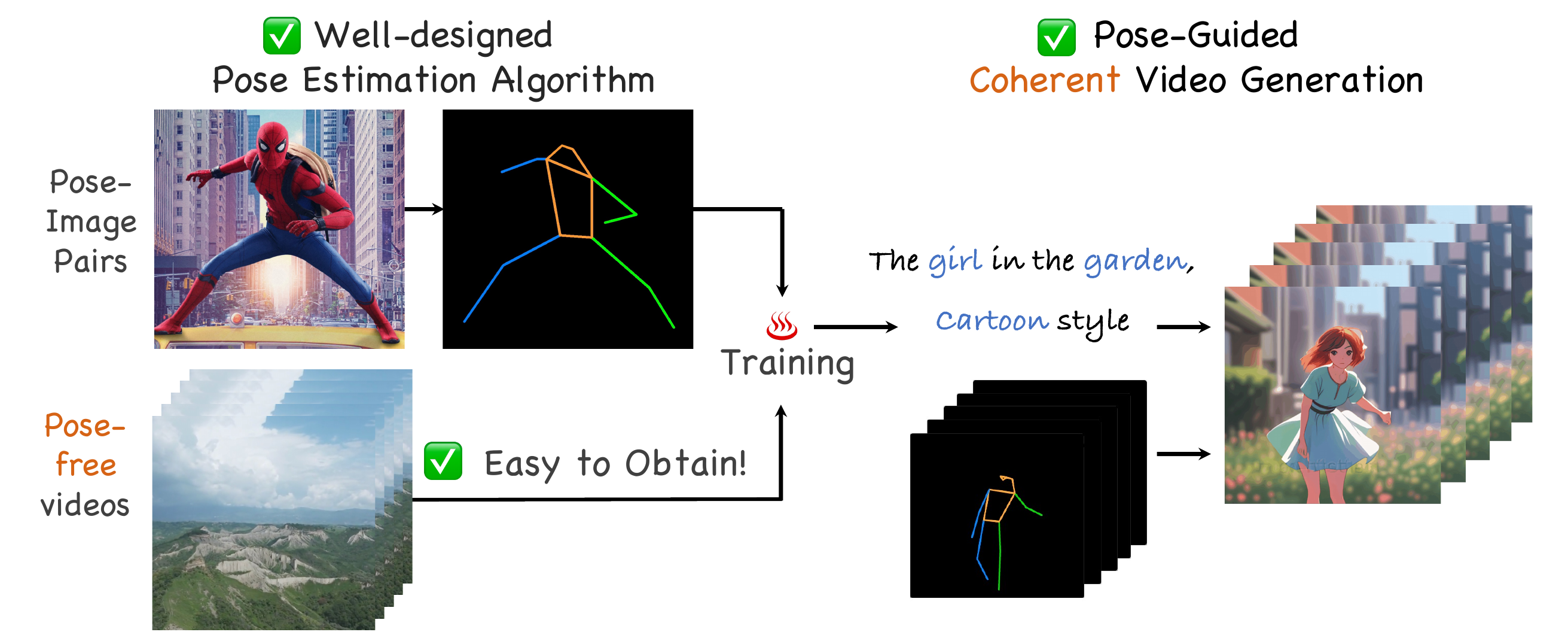}
    \caption{\textit{\textbf{High-level overview of training scheme.}} Images with diverse characters are easy to get from the public large-scale captioned image dataset~\cite{schuhmann2021laion}. We can exploit the off-the-shelf pose estimation algorithm to obtain accurate pose-image pairs. However, captioned video datasets of diverse characters are absent.
    Therefore, we leverage the in-the-wild videos which are easy to obtain to learn temporal coherence.
    Through this separate learning strategy, our tunable blocks also learn to synthesize creative and diverse videos with pose-specific motion while maintaining realistic temporal coherence.
    }
    \label{fig:dataset_explain}
    \vspace{-1em}
\end{figure}

\section{Introduction}
Recent advances in text-to-image~(T2I) synthesis~\cite{ldm, ho2022imagen, ma2023adapedit} have demonstrated impressive performance and creativity, enabling the conversion of textural descriptions into visually compelling creations.
Such AI generative systems excel in various aspects, including rendering realistic and diverse appearances~\cite{mou2023t2i, zhang2023adding, ma2022visual, gao2023backdoor}, powerful editing~\cite{prompt-to-prompt,qi2023fatezero, ma2023lmd}, composition~\cite{blended-latent}, generalization capabilities, and creating new art that satisfies people’s imagination~\cite{mou2023t2i, zhang2023adding}. 
However, in text-to-video generation, there still have limited applications due to the restriction of the high-quality video dataset and the video generative prior model.

\begin{figure*}
    \centering
    \includegraphics[width=0.9\textwidth]{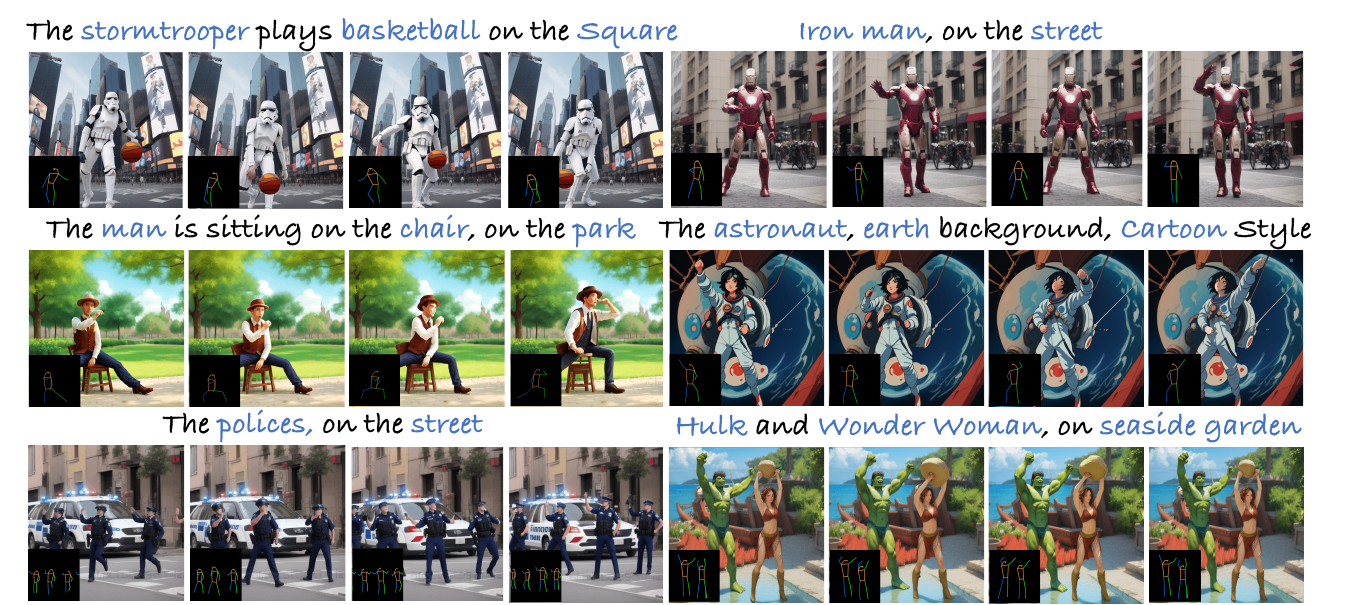}
    \captionof{figure}{
    \textit{\textbf{Pose-Guided Text-to-Video Generation.}}
    We propose an efficient training scheme to empower the ability of the pretrained text-to-image model~(\textit{i.e.}, Stable Diffusion~\cite{ldm}) to generate pose-controllable character videos with minimal data requirements.
    Thanks to the proposed method, we can generate various high-deﬁnition pose-controllable character videos that are well-aligned with the pose sequences and the semantics of text prompts.}
\end{figure*}

In this work, we aim to create high-quality character videos from text descriptions and also control their pose via the given control signal, \textit{i.e.}, human skeletons.
To achieve this, we design an efficient training scheme using easily obtained image datasets and the pre-trained T2I models for controllable text-to-video generation. 
Specifically, given image-pose pairs and pose-free videos, we design a novel two-stage training strategy with carefully tuned blocks through a pretrained text-to-image model, \textit{i.e.}, stable diffusion~\cite{ldm}. 
In the first stage, we add the control ability into the pre-trained text-to-image model via an additional multi-layer pose encoder. In the second stage, we train our model on pose-free videos to learn temporal consistency, such as coherent backgrounds, and avoid flickering. 
To make the pre-trained T2I model suitable for video inputs, we make several key modifications.
Firstly, we add extra temporal self-attention layers~\cite{singer2022make,tuneavideo,qi2023fatezero, ma2023magicstick, ma2023follow, he2023strategic, li2023finedance} to the stable diffusion network.
Secondly, inspired by the recent one-shot video generation model~\cite{tuneavideo}, we reshape the attention to cross-frame attention for better content consistency. 
We then fine-tune the video model for text-to-video generation tasks without the pose control. 
During the training of the second stage, we only update the parameters related to the temporal consistency, \textit{i.e.}, the newly introduced temporal self-attention and the cross-frame attention blocks, while fixing other parameters. 
Since the pose control is ensured by our pose encoder and the temporal information is learned via the video dataset and well-designed temporal modules, after the training of two stages, our method manages to generate pose-controllable character videos with text-editable appearance.

Overall, our proposed method is equipped with delicate designs to generate videos that offer flexible control through pose sequence and textual descriptions. 
Moreover, our model inherits the robust concept generation and composition capabilities of the pre-trained T2I model. As a result, it manages to generate diverse characters, backgrounds, appealing styles, and multiple characters.
Although our main focus is on pose control, our approach can be easily extended to other control modalities~(\textit{i.e.}, depth, edges, sketch, \textit{e.g.})  as demonstrated in concurrent works including T2I adapter~\cite{mou2023t2i}, and ControlNet~\cite{zhang2023adding}.


In summary, our contributions are:
\begin{itemize}
\item We tackle a new task of pose-controllable text-to-video generation, and introduce the LAION-Pose dataset to facilitate the learning of pose-content alignment.
\item To overcome the challenge of lacking diverse character videos, we decouple the task into two subproblems: pose alignment and temporal coherence. We then propose a two-stage training mechanism by carefully tuning different sets of parameters in different datasets.
\item Extensive experiments compared with various baselines demonstrate the superiority of our approach, in terms of generation quality, text-video alignment, pose-video alignment, and temporal coherence.
\end{itemize}



\begin{figure*}
    \centering
    \includegraphics[width=0.95\textwidth]{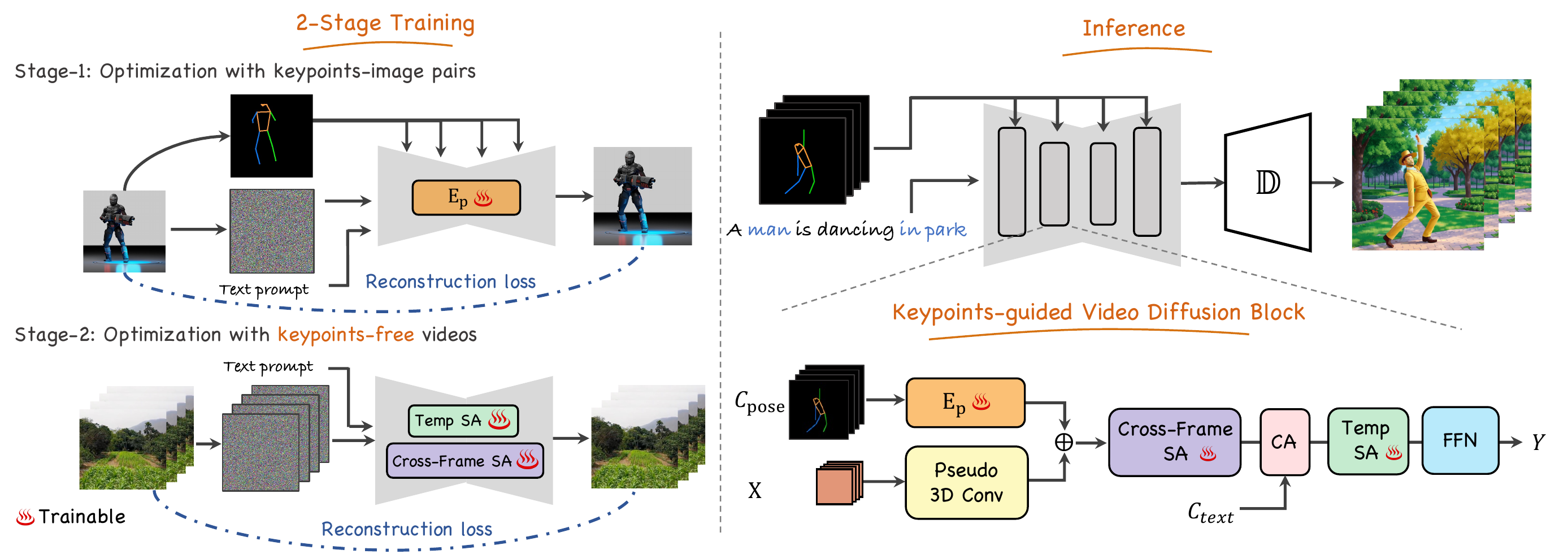}
    \vspace{-1em}
    \caption{\textit{\textbf{Overview.}} We propose a two-stage training strategy to effectively learn image-pose alignment from our proposed LAION-Pose dataset and learn temporal coherence from natural videos without pose annotations.
    During the first-stage training, only the pose encoder $E_p$ is trainable to learn the pose control.
    During the second-stage training, only the temporal modules are trainable, including the temporal self-attention (SA) and the cross-frame self-attention.
    During inference, temporally coherent videos are generated by giving a text describing the target character and corresponding appearance and a pose sequence representing the motion.
    Most parameters from pre-trained stable diffusion are frozen, including the pseudo-3D convolution layers and the cross-attention (CA) and feed-forward network~(FFN) modules.
    }
    \label{fig:pipeline}
    \vspace{-1em}
\end{figure*}

\section{Related Work}

\subsection{Text-to-Video Generation}
Synthesizing natural videos is a challenging task due to the complex and high-dimensional structural characteristics involved. Early works focus on Generative adversarial networks~(GAN~\cite{gan}), which generate samples from the Gaussian distribution via a two-player minimax game. However, GAN-based frameworks are not stale to train and might be hard to model the large-scale dataset. Thanks to the ability of large language models~\cite{clip} and transformer~\cite{transformer}, more current works generate video from the text description. \textit{e.g.}, GODIVA~\cite{wu2021godiva} extends VQ-VAE \cite{oord2018neural} to text-to-video generation by mapping text tokens to video tokens. N\"{U}WA~\cite{wu2021nuwa} proposes an auto-regressive framework that can be used for both text-to-image and video generation tasks. 
CogVideo~\cite{hong2022cogvideo} improves video generation quality by extending CogView-2~\cite{ding2022cogview2} to T2V generation through the incorporation of temporal attention modules and pre-trained text-to-image models. As for the popularity of the diffusion-based method,
Video Diffusion Models~(VDM)~\cite{ho2022video} utilizes a factorized space-time U-Net to directly perform the diffusion process on pixels.
Imagen Video~\cite{ho2022imagen} enhances VDM by implementing cascaded diffusion models and $v$-prediction parameterization.
Similar to CogVideo, Make-A-Video~\cite{singer2022make} extends the significant progress made in the diffusion-based model in text-to-image generation~\cite{ldm} to the T2V task. 
Similar to Make-A-Video~\cite{singer2022make}, we extend image synthesis diffusion models to video generation by introducing temporal connections into a pre-existing image model. Differently, we decouple the motion and appearance in a way that allows for a controllable generation.

\subsection{Pose-to-Video Generation} Generating the realistic human video from the driving signal, \textit{e.g.}, key points, has been studied extensively during recent years, especially with the unprecedented success of GAN-based \cite{mirza2014conditional} models for conditional video synthesis.
Vid2vid~\cite{vid2vid} presents a video synthesis approach that utilizes conditional GANs, incorporating optical flow and temporal consistency constraints along with multiple discriminators to generate pose video sequences that are both realistic and diverse. Similar work has also been proposed by \cite{everybodydance}, where they over-fit two  dance videos by transferring the movements from a source person to a target person through keypoints. However, these methods need to train for each pair. To solve this problem, fewshot-vid2vid~\cite{fewshotvid2vid} extends vid2vid to few-shot settings, where the new image can be animated through the learned model directly or a few-shot adaption. LiquidwarpingGAN~\cite{lwb} proposes a unified framework for human motion imitation using liquid warping and adversarial learning. 
FOMM~\cite{fomm} has designed a first-order motion representation method that is capable of driving any single image through well-decoupled motion fields.
FRAA~\cite{mraa} extends FOMM by a novel motion representation and synthesis framework for articulated characters using pose-dependent motion embedding. 
Nevertheless, while these methods are capable of generating or driving images that resemble the original data distribution, they fall short in their ability to effectively edit videos through text.

\subsection{Controllable Diffusion Models} Due to the ambiguity in generating videos or images from a text description, there has been an increasing focus on conditional image synthesis. 
Early work generates controllable results by image editing method~\cite{sdedit} directly.
More recently, several conditional strategies have been proposed~\cite{mou2023t2i, zhang2023adding}. 
These methods train an additional encoder or adaptor to map the extracted information, \textit{e.g.}, depth map, keypoints, and segmentation map to controlled features. However, since their methods are still image models, the results suffer from the flickering issue when applying these methods to video.


\section{Method}
Our text-guided pose controllable video generation task aims to generate realistic videos from pose sequences and the appearance description from the text. Our method is based on the pretrained text-to-image model, \textit{i.e.}, latent-diffusion model~\cite{ldm} where we make several modifications suitable to our task. In this section, we first give a brief introduction to the latent diffusion model in Sec.~\ref{sec:latent}. Then, we show the details of our pose-guided controllable text-to-video generation network in Sec.~\ref{sec:pose2video}.

\begin{figure*}
    \centering
    \includegraphics[width=0.8\textwidth]{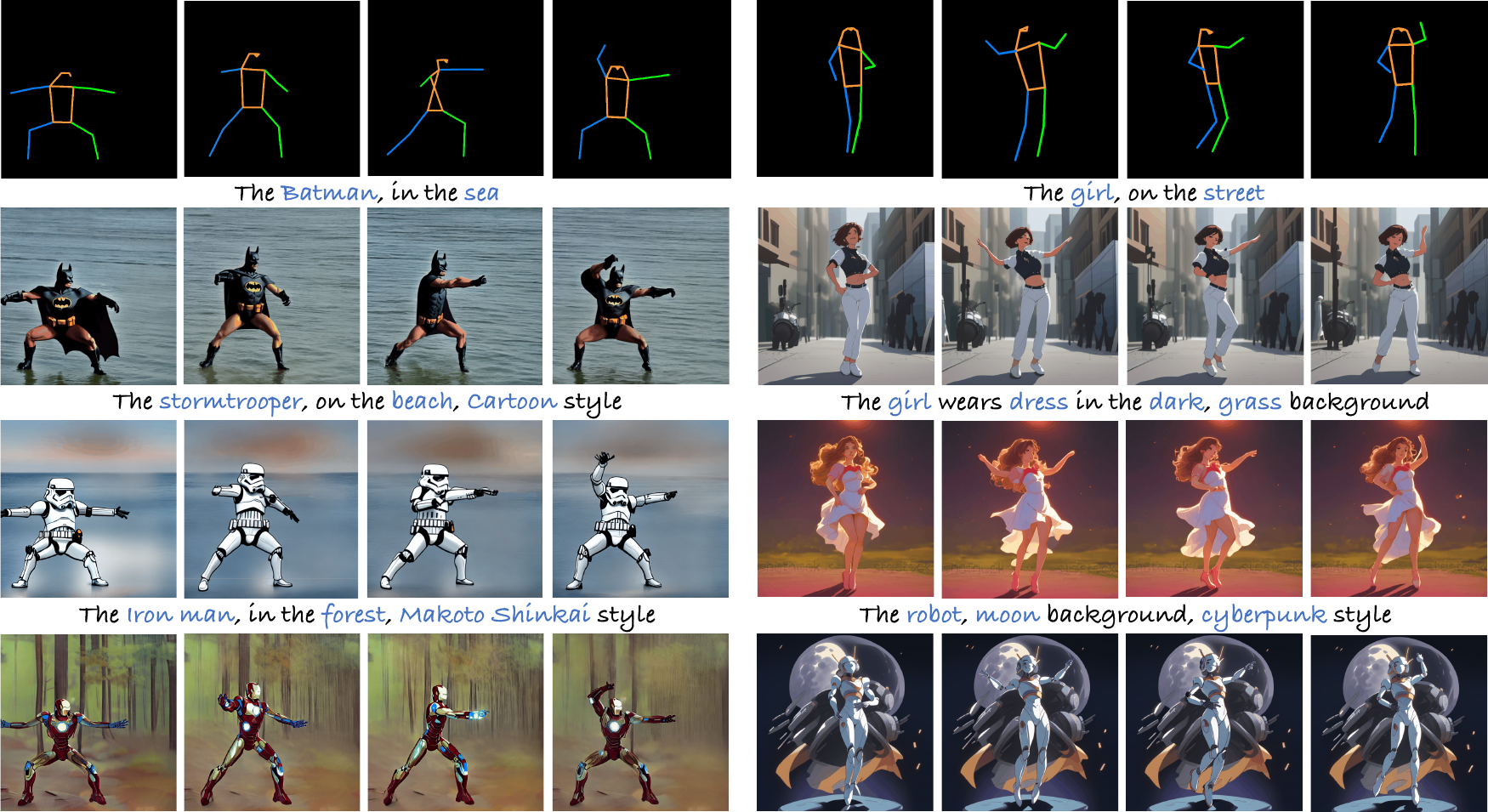}
    \caption{
    \my{
    \textit{\textbf{Results regarding various pose sequences and text prompts.}} Our method can generate videos with high content diversity and temporal coherence, following the pose guidance and semantics of text prompts.
    }
    }
    \label{fig:application}
    \vspace{-1em}
\end{figure*}

\subsection{Preliminary: Latent Diffusion Models}
\label{sec:latent}
Latent diffusion models~(LDM)~\cite{ldm} is a type of diffusion model that models the distribution of the latent space of images and has shown remarkable image synthesis performance recently.
It consists of two models: an autoencoder and a diffusion model.
The autoencoder learns to reconstruct the images via an encoder $\encoder$ and a decoder $\decoder$. The encoder firstly projects the image $\bx$ to a lower dimensional latent: $\bz=\encoder(\bx)$, and the decoder reconstructs the original image from the latent: $\tilde{\bx} = \decoder(\bz)$. 

The diffusion model learns the distribution of image latent $\bz_0 \sim p_{data}(\bz_0)$ via DDPM~\cite{ddpm} and generates new samples in the latent space.
The generation procedure is a gradual backward denoising process with $T$ timesteps starting from pure gaussian noise $\bz_t$ to a novel sample $\bz_0$:
\begin{align}
p_\theta(\latent_{0:T}) & \coloneqq p(\latent_T) \prod_{t=1}^{T} p_\theta(\latent_{t-1}|\latent_t), \\
p_\theta(\latent_{t-1}|\latent_t) & \coloneqq \mathcal{N}(\latent_{t_1}; \mu_\theta(\latent_t, t), \Sigma_\theta(\latent_t, t)),
\label{eq:backwardprocess}
\end{align}
The Markov chain is a gradually forward noising process via a predefined noise schedule $\beta_1, \dotsc, \beta_T$:
\begin{align}
q(\latent_{1:T} | \latent_0) & \coloneqq \prod_{t=1}^T q(\latent_t | \latent_{t-1} ),\\
q(\latent_t|\latent_{t-1}) & \coloneqq \mathcal{N}(\latent_t;\sqrt{1-\beta_t}\latent_{t-1},\beta_t \bI). \label{eq:forwardprocess}
\end{align}
At each timestep, a random noise $\bepsilon$ is drawn from a diagonal Gaussian distribution, and a time-conditioned denoising model $\theta$ is trained to predict the added noise in each timestep with a simple MSE error:
\begin{align}
 \mathcal{L}_\mathrm{simple}(\theta) \defeq \left\| \bepsilon_\theta(\latent_t, t) - \bepsilon \right\|_2^2 \label{eq:training_objective_simple},
\end{align}
\subsection{Pose-guided Text-to-Video Generation}
\label{sec:pose2video}

Due to the scarcity of qualified video-pose pairs in various datasets, we decide to decouple temporal and control conditions, whereas our model learns the pose control capability from images and the temporal consistency from videos. Therefore, we train our model in two different stages. As shown in Fig.~\ref{fig:pipeline}, our method contains two different stages of training for different purposes. Below, we first briefly introduce the denoising network of the Latent Diffusion. Then, we give the details of each stage.

\noindent\textbf{Base Model Architecture.}
The widely-used diffusion model~\cite{ldm} for image synthesis employs U-Net~\cite{ronneberger2015u} for denoising, which is a multi-stage neural network architecture that involves spatial downsampling followed by an upsampling. 
Each stage of the U-Net~\cite{ronneberger2015u} consists of several attention blocks and layers of 2D convolutional residual blocks.
The attention block is constructed from a spatial self-attention, a cross-attention, and a feed-forward network~(FFN). 
The spatial self-attention is utilized for similar correlation by the locations of the latent in representation, while the cross-attention considers the correspondence between latent and conditional inputs~(such as text).

\noindent\textbf{Training Stage 1: Pose-Controllable Text-to-Image Generation.}
In this stage, we train the pose-controllable text-to-image models.
\my{However, the current method does not have pose-image-caption pair} for pose generation. We, therefore, collect the human skeleton images in the LAION~\cite{schuhmann2021laion} by MMpose~\cite{mmpose2020}, only retaining images that could be detected more than 50\% of the key points. Finally, an image-text-pose dataset named LAION-Pose is formed. 
This dataset contains diverse human-like characters with various background contexts.
To incorporate pose conditions into the pre-trained text-to-image denoising model, we fix all the parameters in the original denoising U-Net and propose a simple and lightweight approach that utilizes multiple 3D convolutional layers as the pose encoder \yq{and insert them into each block of U-Net}. 
In detail, we use this convolutional layer \yq{in each block} to extract pose features from the input pose sequence. After that, \my{the
pose features are downsampled to different resolutions (\textit{i.e.} $64 \times 64$, $32 \times 32$, $16 \times 16$, $8 \times 8$).}
We inject this additional controlling information into pre-trained U-Net via a residual connection, by adding the feature into \yq{each} layer of the U-Net model as shown in Fig~\ref{fig:pipeline}.
This residual injection has the advantage of preserving the generation ability of the pre-trained diffusion model, allowing us to update a few of the parameters to add extra pose-controlling information.
Additionally, we can easily drop the condition information by setting the residual to all zero to restore the original diffusion model as demonstrated in Fig.~\ref{eq:training_objective_simple}.
\my{Meanwhile, we also try different ways to incorporate the pose condition into diffusion, such as concatenating it into the channel dimension of the input video,} however, the performance is not good as ours as shown in the experiments. 
We use the simple noise reconstruction loss as the original stable diffusion to tune the injected parameters.

\noindent\textbf{Training Stage 2: Video Generation via Pose-free Videos.} However, the stage 1 model can generate similar pose videos yet the background is inconsistent. Thus, we further finetune the model from our first stage on the pose-free video dataset HDVLIA~\cite{xue2022advancing}. This dataset contains continuous in-the-wild video text pairs. To generate temporal consistent video, we utilize the generative prior of the text-to-image model for our pose-to-video generation, following previous Video Diffusion Model baselines~\cite{ho2022imagen, he2022latent, ho2022video}, we inflate first convolution layer of the pretrained U-Net to a $1 \times 3 \times 3$ convolution kernels as a pseudo-3D convolutional layer for video inputs and appending temporal self-attention for temporal modeling \cite{Carreira2017}. 
To further maintain the temporal consistency, we leverage the cross-frame self-attention~(SA)~\cite{tuneavideo} between frames. Differently, we make it to generate longer video sequences by simply reusing the noise from each time step in the previous sampling process of DDIM. 
In detail, assuming that $T$ frames are sampled each time, we add the noise of the last $\frac{T}{2}$ frames to the next loop as prior knowledge after the first sampling.
Note that, throughout the denoising process, the noise at each time step is added to the prior knowledge to ensure the temporal consistency of the generated long videos. As shown in Fig.~\ref{fig:pipeline}, the proposed method only tune the cross-frame self-attention and the temporal self-attention for video generation.

\noindent\textbf{Discussion.} Empowered by the above two-stage training with carefully-designed tune-able blocks for each stage, the proposed method can generate continuous pose-controllable video from the easily obtained datasets, \textit{e.g.}, the image pairs of the human and pose and the random video. Our method can also be used in other related conditional video generation tasks as shown in concurrent conditional image generation works~\cite{zhang2023adding, mou2023t2i}.

\section{Experiments}

\subsection{Implementation Details}

We implement our method based on the official codebase of Stable Diffusion~\cite{ldm}\footnote{https://github.com/CompVis/stable-diffusion.}
and
the publicly available 1.4 billion parameter T2I model\footnote{https://huggingface.co/CompVis/stable-diffusion-v1-4.} and several LoRA models from CivitAI\footnote{https://civitai.com/}.  We freeze the image autoencoder to encode each video frame to latent representation individually.
We first train our model for 100k steps on the Laion-Pose.
Then,
we train for 50k steps on the HDVILA\cite{xue2022advancing}.  
The eight consecutive frames at the resolution of $512 \times 512$ from the input video are sampled for temporal consistency learning.
The training process is performed on 8 NVIDIA Tesla 40G-A100 GPUs 
and
can be completed within two days.
At inference, we apply DDIM sampler~\cite{song2020denoising} with classiﬁer-free guidance~\cite{ho2022classifier} in our experiments for Pose-guided T2V generation.

\begin{figure}[t]
  \centering
  \includegraphics[width=0.8\columnwidth]{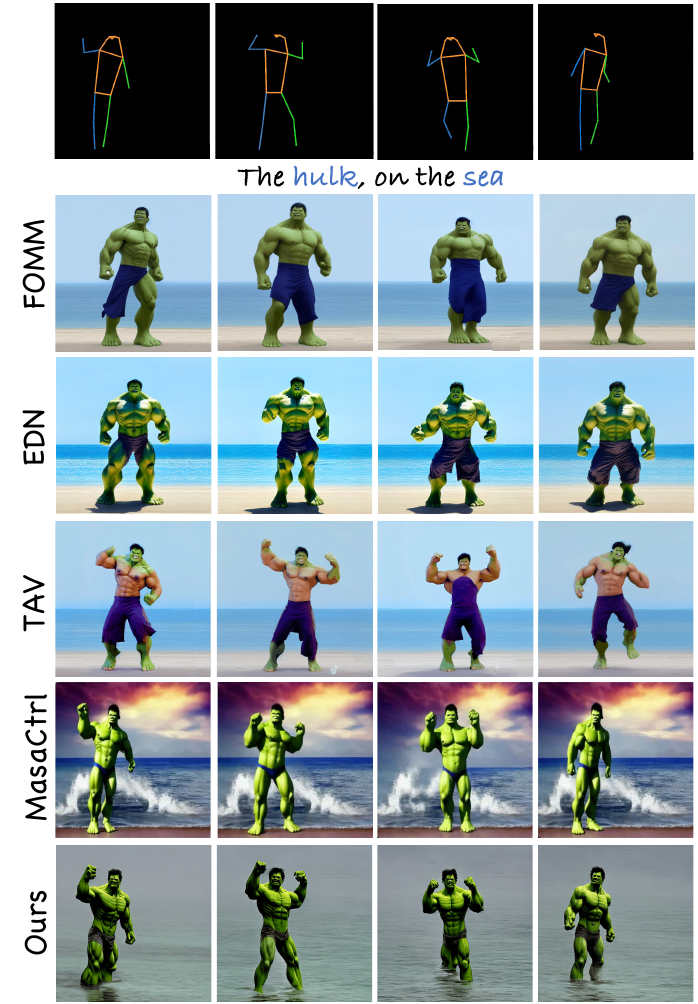}
  \caption{\textit{\textbf{Qualitative comparison between evaluated methods}}.
For FOMM~\cite{fomm} and EDN~\cite{everybodydance}, we generate results by driving one image using source video. As for TAV~\cite{tuneavideo}, it learns the motion from source video. In masactrl~\cite{cao2023masactrl}, we leverage T2I-adapter to control the motion effectively. Our approach achieves consistent background and appearance in the generated video clip.
  }
  \label{fig:compare_4}
  \vspace{-0.5cm}
\end{figure}

\begin{figure*}
\vspace{-0.7cm}
    \centering
    \includegraphics[width=\textwidth]{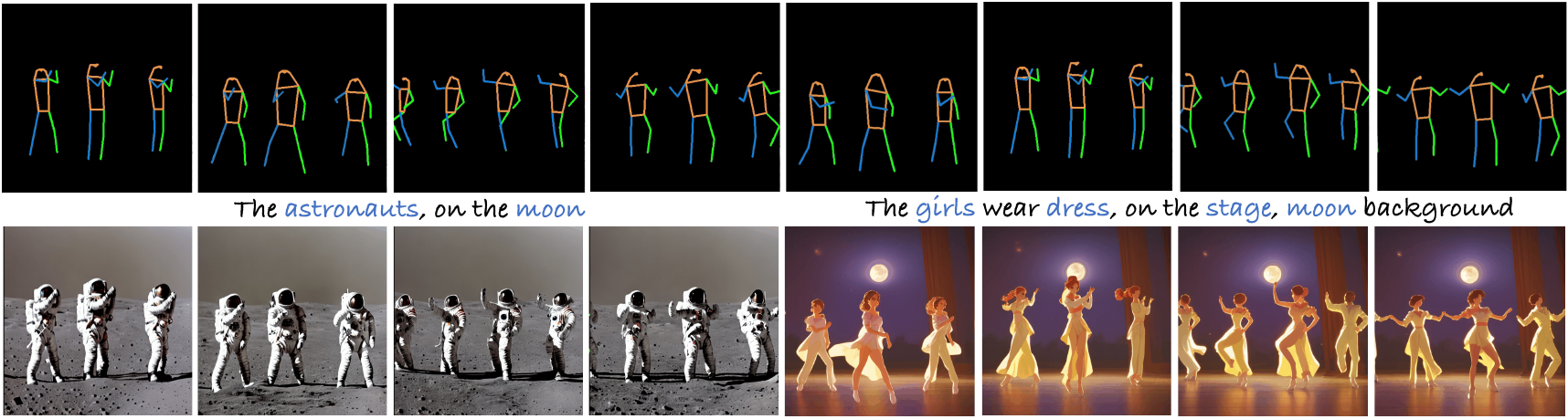}
    \caption{\textit{\textbf{Results regarding multiple skeletons.}}
    Given a sequence of poses of multiple persons, our approach can also generalize well \my{in} this situation. 
    Furthermore, as shown in the second row, the character identity can be specified via the input textual descriptions such as \my{\enquote{the astronauts on the moon}}.
    Through this application, we can bring irrelevant characters into one video while specifying the background context.
    }
    \label{fig:multiple}

\end{figure*}

\subsection{Applications} 

We provide several applications of our approach in generating videos with novel visual concepts (e.g., character, backgrounds, styles, etc.)  and pose guided by the text prompt. 
Here, 
we showcase two various pose-guided training examples 
and
their variations with edited prompts. 
More examples can be found in the supplementary material.

\noindent\textbf{Diverse character generation.} 
One of the applications of our method is to edit the character with diverse appearances in the generated video. 
\my{From the Fig.~\ref{fig:application},}
we can see that our approach could generate videos with customized characters by changing the corresponding descriptions in the input text prompt.
For example,
given an input pose sequence,
we manage to generate \enquote{batman}, \enquote{stormtrooper} and \enquote{Iron man} etc., with the same motion by varying the input text prompt.
The background and the character's appearance successfully keep consistent in different video frames.

%


\noindent\textbf{Diverse background generation.}
Our approach also enables generate the natural video background (\textit{i.e.}., the place where the character is), 
while keeping the pose of the character and the temporal information consistent. 
For example, 
we can synthesize the \enquote{sea} and \enquote{beach} background (the left 2nd and 3rd rows of Fig.~\ref{fig:application}). Meanwhile, we manage to generate special background such as \enquote{moon} and \enquote{grass} (the right 3rd and 4th rows of Fig.~\ref{fig:application}). The generated video maintains a high level of background consistency and effectively matches the pose and text prompts.




\noindent\textbf{Stylized character video generation.} 
Thanks to the broad knowledge of per-trained T2I models and LoRA~\cite{hu2022lora},
our method is able to generate videos with different impressive styles.
By appending the style description (\textit{e.g.}, cartoon style, makoto shinkai style, or cyberpunk style) to the sentence,
our method produces videos in that style with consistent \my{poses} and semantics (the left 2nd and 3rd rows, the right 3rd row of Fig.~\ref{fig:application}).

\noindent\textbf{Multiple-characters generation.}
We found that our approach even enables translating multiple skeletons to video,
(\textit{e.g.}, the astronauts and the girls
in the 2nd row of Fig.~\ref{fig:multiple}).
\my{
We note that the properties of generated video could nicely match input pose sentences and text prompts, indicating our \modelname powerful generalization ability.}

\begin{figure}[t]
\vspace{-0.6cm}
  \centering
  \includegraphics[width=0.95\columnwidth]{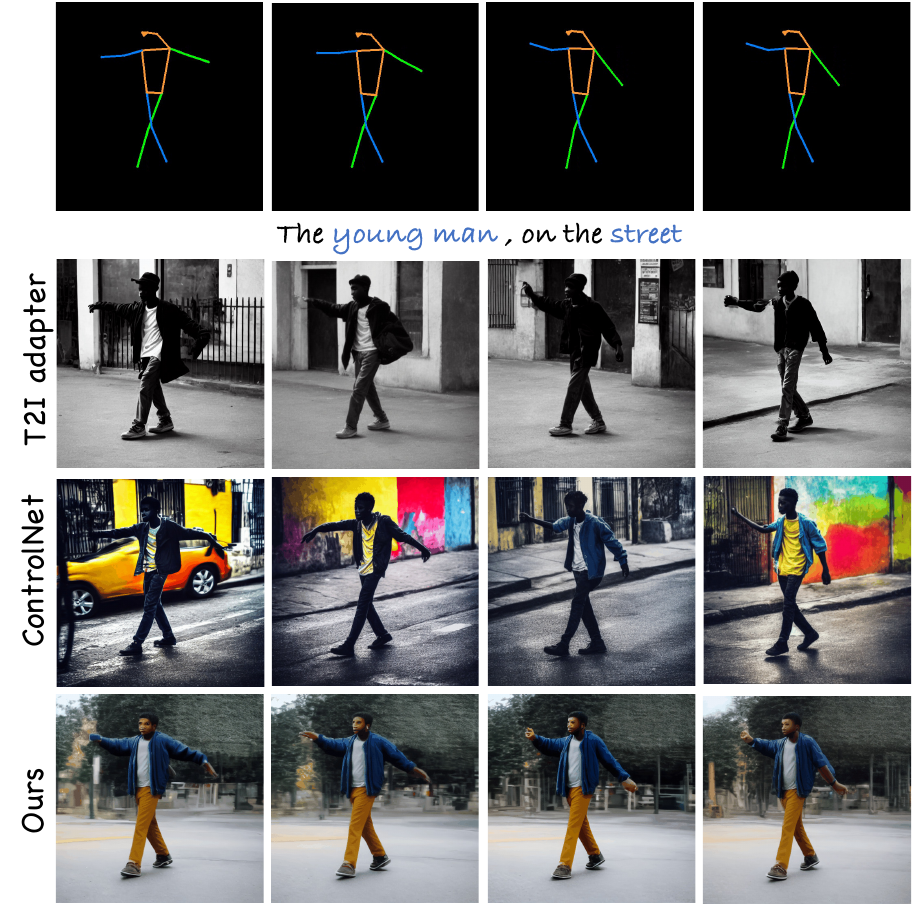}
  \caption{\textit{\textbf{Comparison with ControlNet~\cite{zhang2023adding} and T2I adapter~\cite{mou2023t2i}}}. These two approaches can translate poses into realistic images. However, they tend to generate different character appearances with different backgrounds.
  Our approach achieves consistent appearance in the generated video clip.
  }
  \label{fig:controlNet}
  
\end{figure}

\subsection{Comparison with Baselines}
We provide quantitative and qualitative comparisons with two concurrent works, including Tune-A-Video \cite{tuneavideo}, and
ControlNet \cite{mirza2014conditional}. 
ControlNet is based on a pre-trained Stable Diffusion~\cite{he2022latent} and
mainly applied for various conditional image generation.
\my{It is extensively trained on a large-scale dataset of 200k pose-image-caption pairs collected from the internet.}

\noindent\textbf{Quantitative results.}
1) \emph{CLIP score}: 
We follow~\cite{ho2022imagen} to evaluate our \modelname on CLIP score \cite{nguyen2021clipscore, park2021benchmark} or video-text alignment.
We compute \my{CLIP score} for each frame and then average them across all frames. 
The ﬁnal CLIP score is calculated on 1024 video samples. 
The results are reported in Tab.~\ref{tab:qr}. 
Our \modelname produces a higher CLIP score, demonstrating better video-text alignment than the other two approaches. 
2) \emph{Quality}: Following Make-A-Video~\cite{singer2022make}, We conduct \my{the human evaluation of videos' quality across a test set containing 32 videos. In detail,} we display three videos in random order and request the evaluators to identify the one with superior quality. Our observations indicate that the raters favor the videos generated by our \modelname over Tune-A-Video and ControlNet in video quality.  
3) \emph{Pose Accuracy}: 
we regard the input pose sentence as ground truth video and evaluate the average precision of the skeleton on 1024 video samples. 
For a fair comparison, we adopt the same pose detector~\cite{sun2019hrnet} for both the processing of LAION-Pose collecting and evaluation.
The results show that our model achieves comparable performance with ControlNet.
4) \emph{Frame Consistency}:
Following \cite{esser2023structure},
we report frame consistency measured via CLIP cosine similarity of consecutive frames, and the results are shown in Tab.~\ref{tab:qr}.
\my{Our model outperforms the ControlNet regarding temporal consistency, which demonstrates the necessity of our temporal designs.}
In addition, we obtain a comparable score with Tune-A-Video. However, Tune-A-Video relies on an input video and needs to overfit the input video, which is hard to sever as a general video generation model.

\noindent\textbf{Qualitative results.}
We first compare our \modelname with other four methods with the same pose sequences and text prompt in Fig.~\ref{fig:compare_4}. Our \modelname obtains the better performance in consistency and artistry. In similar setting, We also compare our \modelname and ControlNet and T2I-adapter in Fig.~\ref{fig:controlNet}.
It is apparent to discover that there is an inconsistent background in ControlNet (\textit{e.g.}, the color of the street and the shirt of the boy).
This phenomenon is common and can also be encountered in the 3rd row of Fig.~\ref{fig:controlNet} (the wall color). 
In contrast, our \modelname could effectively address the issue of temporal consistency and learn good inter-frame coherence.


\begin{table}[t]

\resizebox{0.98\columnwidth}{!}{
  \begin{tabular}{lcccc}
    \toprule
    Method & CS & QU (\%) & PA (\%) & FC (\%)\\
    \midrule
    FOMM~\cite{fomm} & 22.93 & 0.8 & 11.7 & {81.25}\\
    Everybody dance now~\cite{everybodydance} & 23.04 & 1.3 & 13.7 & {79.83}\\
    Tune-A-Video~\cite{tuneavideo} & 23.57 & 23.81 & 27.74 & \textbf{93.78}\\
    ControlNet~\cite{zhang2023adding} & 22.31 & 6.69 & 33.23 & 54.35 \\
    T2I adapter~\cite{mou2023t2i} & 22.42 & 8.27 & 33.47 & 53.86 \\
    Masactrl~\cite{cao2023masactrl} & 23.64 & 19.17 & 33.19 & {87.64}\\
    Ours & \textbf{24.09} & \textbf{39.96} & \textbf{34.92} & 93.36 \\
    \bottomrule
  \end{tabular}
 }
\caption{
    \textit{\textbf{Quantitative comparisons with related works.}}
    CS, QU, PA, and FC represent clip score, quality, pose accuracy, and frame consistency, respectively.
    }
\label{tab:qr}
\end{table}

\subsection{Ablation Study}

\begin{figure}[t]
\vspace{-0.6cm}
  \centering
  \includegraphics[width=0.95\columnwidth]{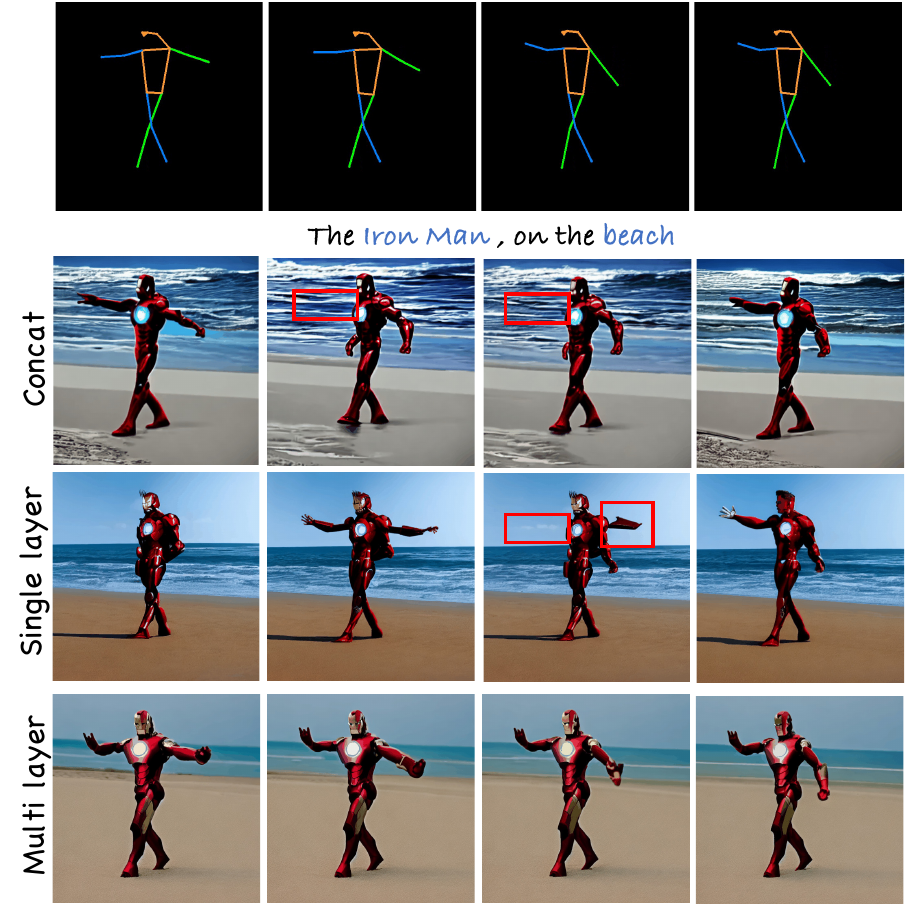}
  \caption{\textit{\textbf{Ablation study.}} Including the condition injection mechanism (feature residual and channel-wise concatenation) and the number of layers with conditional signals.
  }
  \label{fig:ablation}
\vspace{-0.1cm}
\end{figure}

\noindent\textbf{Effect of residual pose encoder.}
We ablate the feature residual design of the proposed pose encoder.
To add controlling information to the diffusion model, one natural way is to directly concatenate the condition onto the input noisy latent of the model~\cite{ldm}. However, we discover that the concatenation approach yields worse performance than the residual approach.
We compare the generation results of residual and concatenation approaches for injecting the pose information in the first row and the second row of Fig.~\ref{fig:ablation}.
We discover that the feature residual manner for adding extra controls can preserve more of the generation ability of pre-trained stable diffusion. 
This is because the concatenation mechanism needs to retrain the first convolutional layer to meet the number of input channels which sacrifices the pretrained high-quality image synthesis prior. 

\noindent\textbf{Number of layers of condition control.}
We also ablate the number of layers injecting the controlling signals in order to achieve better controllability (see comparisons between the second row and the third row in Fig. \ref{fig:ablation}).
Our results indicate that adding controls to more layers can lead to improved \my{pose-frame} alignment.
Specifically, adding the pose into one single layer results in the mismatch between the target arm and the generated arm of \my{Iron man.}

\section{Conclusion}
In this paper, we tackle the problem of generating text-editable and pose-controllable character videos. To achieve this, we reform and tune from the pre-trained text-to-image model due to its powerful semantic editing and composition capacities. We thus design a new two-stage training scheme that can utilize the large-scale image pose pairs and the diverse pose-free dataset.
In detail, in the first training stage, we use a pose encoder to inject the pose information into the network structure and learn from the image-pose pair for the pose-controllable text-to-image generation. In the second training stage, we inflate the image model to a 3D network to learn the temporal coherence from the pose-free video.
Equipped with our several new designs, we can generate novel and creative temporally coherent videos with the preservation of the conceptual combination ability of the original T2I model. 

\section{Acknowledgments}
This project was supported by the National Key R$\&$D Program of China under grant number 2022ZD0161501, Shenzhen Key Laboratory of next generation interactive media innovative technology(No:ZDSYS20210623092001004).

\bigskip
\bibliography{aaai24}

\end{document}